\documentclass{article}
\usepackage{spconf,amsmath,graphicx}
\usepackage{nccmath}
\usepackage{graphicx}
\usepackage{mathrsfs}  
\usepackage{amsthm,lipsum}
\usepackage{algorithm}
\usepackage{algorithmic}
\usepackage{subcaption}

\usepackage{hyperref,graphicx,amsmath,amsfonts,amssymb,bm,url,breakurl,epsfig,epsf,color,MnSymbol,mathbbol,fmtcount,semtrans,caption,subcaption,multirow,comment, boldline, enumitem}
\usepackage{wrapfig}
\usepackage{amssymb}

\makeatletter
\providecommand*{\boxast}{%
  \mathbin{% as \boxplus and \boxtimes
    \mathpalette\@boxit{*}%
  }%
}
\newcommand*{\@boxit}[2]{%
  \sbox0{$\m@th#1\Box$}%
  \ifx#1\displaystyle \ht0=\dimexpr\ht0+.05ex\relax \fi
  \ifx#1\textstyle \ht0=\dimexpr\ht0+.05ex\relax \fi
  \ifx#1\scriptstyle \ht0=\dimexpr\ht0+.04ex\relax \fi
  \ifx#1\scriptscriptstyle \ht0=\dimexpr\ht0+.065ex\relax \fi
  \sbox2{$#1\vcenter{}$}% \ht2 is positionn of math axis
  \rlap{%
    \hbox to \wd0{%
      \hfill
      \raisebox{%
        \dimexpr.5\dimexpr\ht0+\dp0\relax-\ht2\relax
      }{$\m@th#1#2$}%
      \hfill
    }%
  }%
  \Box
}
\makeatother

  \makeatletter
\def\BState{\State\hskip-\ALG@thistlm}
\makeatother

  \usepackage{mathtools}

\usepackage{tikz}
\usepackage{pgfplots}
\usetikzlibrary{pgfplots.groupplots}

\usepackage{movie15}

\usepackage[bottom,hang,flushmargin]{footmisc} 

\newcommand{\tsn}[1]{{\left\vert\kern-0.25ex\left\vert\kern-0.25ex\left\vert #1 
    \right\vert\kern-0.25ex\right\vert\kern-0.25ex\right\vert}}

\definecolor{darkred}{RGB}{150,0,0}
\definecolor{darkgreen}{RGB}{0,150,0}
\definecolor{darkblue}{RGB}{0,0,200}
\hypersetup{colorlinks=true, linkcolor=darkred, citecolor=darkgreen, urlcolor=darkblue}

\newtheorem{theorem}{Theorem}[section]

\newtheorem{assumption}{Assumption}

\newtheorem{lemma}[theorem]{Lemma}

\newtheorem{definition}[theorem]{Definition}

\newtheorem{remark}[subsection]{Remark}

\newcommand{\distas}{\overset{\text{i.i.d.}}{\sim}}

\newcommand{\beq}{\begin{equation}}

\newcommand{\eeq}{\end{equation}}

\newcommand{\la}{\lambda}

\newcommand{\bt}{\boldsymbol{\theta}}

\newcommand{\Ub}{{\mtx{U}}}

\newcommand{\Pb}{{\mtx{P}}}

\newcommand{\Qb}{{\mtx{Q}}}

\newcommand{\La}{{\boldsymbol{{\Lambda}}}}

\newcommand{\bSi}{{\boldsymbol{{\Sigma}}}}

\newcommand{\Iden}{{\mtx{I}}}
\newcommand{\M}{{\mtx{M}}}

\newcommand{\Sc}{\mathcal{S}}

\newcommand{\Nn}{\mathcal{N}}

\newcommand{\vb}{\vct{v}}

\newcommand{\Fc}{\mathcal{F}}

\newcommand{\x}{\vct{x}}

\newcommand{\W}{\mtx{W}}

\definecolor{emmanuel}{RGB}{255,127,0}

\newcommand{\R}{\mathbb{R}}

\newcommand{\vct}[1]{\bm{#1}}
\newcommand{\mtx}[1]{\bm{#1}}

\numberwithin{equation}{section} 

\def \endprf{\hfill {\vrule height6pt width6pt depth0pt}\medskip}
\newcommand\cL{\mathcal{L}}
\newcommand\cLoss{\mathrm{loss}}

\newcommand\bz{\boldsymbol{z}}

\def\bM{\mtx{M}}

\newcommand{\argmin}{\mathop{\mathrm{argmin}}}
\newcommand{\cLCE}{\cL}

\def\x{{\mathbf x}}

\newcommand{\som}[1]{}%\marginpar{\color{darkblue}\tiny\ttfamily SO: #1}}
\newcommand{\so}[1]{{\color{darkblue}#1}}
\newcommand{\mf}[1]{}%\marginpar{\color{green}\tiny\ttfamily mf: #1}}%{\color{green}#1}}
\newcommand{\rrr}[1]{\it \color{red}}
\newcommand*{\Scale}[2][4]{\scalebox{#1}{$#2$}}

\newcommand{\Pxy}{\mathscr{P}_{x,y}}
\title{Sample Efficient Subspace-based Representations for Nonlinear Meta-Learning}

\name{Halil Ibrahim Gulluk$^{\star\alpha}$ \qquad Yue Sun$^{\dagger\alpha}$ \qquad Samet Oymak$^{\ddagger}$\qquad Maryam Fazel$^{\dagger}$ \thanks{$^{\alpha}$ Equal contribution.}\thanks{$^{\star}$ Halil Ibrahim Gulluk's work is done during an internship at the University of California, Riverside.}}
\address{$^{\star}$ Bogazici University\\$^{\dagger}$ University of Washington\\$^{\ddagger}$ University of California, Riverside}
\begin{document}
%\ninept
%
\maketitle
\begin{abstract}
Constructing good representations is critical for learning complex tasks in a sample efficient manner. In the context of meta-learning, representations can be constructed from common patterns of previously seen tasks so that a future task can be learned quickly. While recent works show the benefit of subspace-based representations, such results are limited to linear-regression tasks. This work explores a more general class of nonlinear tasks with applications ranging from binary classification, generalized linear models and neural nets. We prove that subspace-based representations can be learned in a sample-efficient manner and provably benefit future tasks in terms of sample complexity. Numerical results verify the theoretical predictions in classification and neural-network regression tasks.

%Representation learning is a critical aspect of the modern deep learning algorithms and enables sample efficiency. 
\end{abstract}
%
%\section{Introduction}
%\label{sec:intro}

%{\color{red} YS: ICASSP says ``Papers may be no longer than 5 pages, including all text, figures, and references, and the 5th page may contain only references'' so at most 1 page reference?}

\setlength{\abovedisplayskip}{6pt}
\setlength{\belowdisplayskip}{6pt}
\setlength{\abovedisplayshortskip}{6pt}
\setlength{\belowdisplayshortskip}{6pt}

\vspace{-0.1cm}
\section{Introduction}\label{s:intro}
Meta-learning (and multi-task learning) has proved to be a efficient when available training data is limited. The central idea is exploiting the information (e.g.~training data) provided by earlier related tasks to quickly adapt a new task using few samples.
This idea has a rich history \cite{caruana1997multitask,baxter2000model} and has shown promise in modern machine learning tasks, e.g., in image classification \cite{deng2009imagenet}, machine translation \cite{bojar2014findings} and reinforcement learning \cite{finn2017model}, all of which may involve numerous tasks to be learned with limited data per task.
% between thousands of English and French sentences

 \mf{where? give citation}
  \mf{should we use the term meta-learning or multitask learning? It's better to pick one for consistency}
Modern deep learning algorithms typically exploit the shared information between tasks by learning useful representations \cite{schmidhuber1987evolutionary, thrun2012learning}. The multi-task system was studied by \cite{caruana1997multitask}, and the idea of meta-learning or transfer learning is investigated empirically in modern machine learning framework, showing that the shared representation benefits for training on the new tasks \cite{bengio2013representation, hospedales2020meta, yosinski2014transferable}.  %\cite{bengio2013representation, hospedales2020meta, yosinski2014transferable, raghu2019rapid}. 
An instructive and well-studied problem for meta-learning is mixed linear regression,
for which efficient algorithms and sample complexity bounds are discussed in 
\cite{zhong2016mixed, li2018learning, chen2020learning}. If the tasks lie on a shared low-dimensional subspace, learning this subspace would serve as an efficient representation which helps reduce the search space for future tasks. Once the search space is low dimensional, in order to get the same accuracy, the amount of data required for training is reduced compared to training over the full parameter space. \cite{lounici2011oracle,cavallanti2010linear,maurer2016benefit} propose sample complexity bounds for representation learning for linear multi-task systems. There are study of mixed linear tasks combined with other structures, such as boolean combination of features \cite{balcan2015efficient}, half-spaces \cite{rish2008closed} and sparse representations \cite{argyriou2008convex}.
% \mf{main point: low-dimensional approximation of the feature space makes learning easier in a new task, with fewer samples. Thus in many ML applications one seeks such a representation. can omit 'learned by mixed linear regression'. rewrite parag.}

% The low dimensional subspace of the features, learned by mixed linear regression, is utilized as the search space for training on new tasks.
% \mf{swap the order with next paragraph} 

The recent papers \cite{kong2020meta,kong2020robust} propose meta-learning procedures that involve dimension reduction, clustering and few-shot learning. Here a low-dimensional task subspace is used as the search space for few-shot learning for the new task. 
Another related approach
\cite{du2020few,tripuraneni2020provable}
sets up a nonconvex optimization problem with matrix factors of appropriate sizes, which captures the low dimensional structure. One can apply gradient descent to this nonconvex problem, and studying its behavior requires a nontrivial landscape analysis of the matrix factorization problem.  

However, existing provable algorithms for representation learning are restricted to linear-regression tasks, whereas typical machine learning tasks involve nonlinearity. This can arise from the use of nonlinear models as well as nonlinear label link function (e.g.~generalized linear models).
%. The setup is restrictive compared with the variant types of nonlinear tasks. 
\mf{what does 'objective' mean here? do you mean there are different tasks to be performed with different goals, with the same data?}
A good example is classification problems 
%which represent much of the machine learning applications 
including computer vision and natural language processing \cite{deng2009imagenet, bojar2014findings}. 
In classification tasks, the model is a map from images/text to labels, and the labels are discrete and not linear with respect to the input (i.e.~logistic link function). Another example is the use of nonlinear models such as deep networks
%, which contain many nonlinear activation functions within their layers. 
The existing results for representation learning for the linear-regression setting cannot be easily extended to the nonlinear case. 

\noindent\emph{Can we learn efficient subspace representations for nonlinear tasks such as generalized linear models and neural nets?}
%If one can learn a representation that help classify across several tasks, this representation can be used in this model can be retrieved and transferred to a classifier for other objectives.
%\som{Redundant sentence: In these datasets, there are images of different types of objectives, and the task is to classify which categories each image comes from.}

We consider a realizable setup where the input data is high-dimensional, the \emph{relevant features} lie in a low dimensional subspace and the labels depend only on the \emph{relevant features}. These assumptions are the same as in the existing literature, however we additionally allow for the scenario where labels are possibly an arbitrary nonlinear function of the relevant features. We make the following contributions.

\noindent$\bullet$ \textbf{Efficient representations for nonlinear tasks:} We show that subspace found via method-of-moments (MOM) leads to a consistent estimate of the ground-truth subspace despite arbitrary task nonlinearities, when the data is normally distributed.  We combine this with non-asymptotic learning results to establish sample complexity bounds for representation learning.

% \som{emphasize generalized linear model?}
\noindent$\bullet$ \textbf{Few-shot learning and Applications:} We specialize our results to practical settings with tasks involving binary classification and neural nets. We theoretically and empirically show that subspace-based representation can greatly improve sample efficiency of future tasks.
% \som{More citations are needed! Consider adding a related work (sub)section!}

%that, can we still retrieve the subspace with the same amount of training data?

%{\color{blue}Contribution here}

\vspace{-0.15cm}
\setlength{\topsep}{0.25em}
\newcommand{\xij}{{\x_{i,j}}}
\newcommand{\xijt}{{\x_{i,j}^\top}}
\newcommand{\yij}{{y_{i,j}}}

\section{Problem Formulation}\label{s:prob}
% {\color{red} YS: a paragraph for notation?}
\vspace{-0.1cm}
The meta-learning setup that will be considered in this work consists of two phases: (i) meta-training: prior tasks are used to learn a good representation and (ii) few-shot learning: the new task is learned with few samples.
%\som{Is meta-train/test standard and correct phrases?} 
%{\mf better not to use the meta-train and meta-test phrases if we don't need to} 
In the meta-training phase, we learn the low dimensional space spanned by parameters. In the few-shot learning phase, we use the subspace to learn the model of a new task ideally with few samples.

In the first phase, there are multiple task vectors to infer from.%, each with its own distribution. 
We consider a realizable model where the input and label is associated via a labeling function. One accesses batches of data, each of whom is collected from a task.
%, however we may not know which task it comes from. 
%We make this setup more precise using the following definitions. 
Below, we denote the ground-truth representation by a matrix $\W\in\R^{r\times d}$ whose row space corresponds to the subspace of interest.

%\som{Use $n_j$ for sample size rather than $t_j$ (at least for training)}
\begin{definition} \label{def:task}
\textbf{Meta-training data.} Fix a matrix $\W\in\R^{r\times d}$ satisfying $\W\W^T=\Iden$. The $j$-th task is associated with function $f^j: \R^r\rightarrow \R$. Given input $\x\in\R^{d}$, the label $y$ is distributed as $p_j(y|\x)=p_j(y|\W\x)$\footnote{In words, the label only depends on the relevant features induced by $\W$.} and the expectation satisfies $\mtx{E}(y) = f^j(\W\x)$.
%{\mf give properties like domain and range of $f$ here, first time it appears.}
Suppose there are $n_j$ samples from the $j$-th task sampled i.i.d.~from this distribution and we denote the dataset $\Sc^j = (\xij, \yij)_{i=1}^{n_j}$. Define the full meta-training dataset to be $\Sc = \bigcup_{j=1}^k\Sc^j$.
\end{definition}
Here, $f^j$ is allowed to be any Lipschitz nonlinear function, i.e., a neural network\footnote{In our theoretical results, we treat $f$ as a general linear function, and in experiments we will use a neural network with a specific structure.}.

\begin{definition} \label{def:bin_class}
\textbf{Binary classification.} Suppose $f^j$ takes values over $[0,1]$,
%{\it\color{red}{Is it [0,1] ?}}. The labels $\yij$ are generated as 
\begin{align*}
    \yij = 
     \begin{cases}
       \text{1,} &{\mbox{with probability } f^j(\W \xij),}\\
       \text{0,} &{\mbox{with probability } 1-f^j(\W\xij).}\\
     \end{cases}
\end{align*}
%Here $f^j(a)\in[0,1]$ for all $a\in[-1,1]$, and the event that $f(a) = 0$ is not true with probability $1$ with respect to Lebesgue measure on $[-1,1]$\footnote{In other words, $f$ is not a constant $0$ function.}.
\end{definition}

\begin{definition}\label{def:GLM}
\noindent\textbf{Generalized linear models (GLM)} (which include logistic/linear regression) can be modeled by choosing $f^j$ to be parameterized by a vector $\bt_j\in\R^r$ and a link function $\phi:\R\rightarrow\R$ as $f^j(\W \xij) := \phi(\bt_j^T\W\xij )$.

\end{definition}
%for $\bt_j\in\R^r$, $\phi$ is a function mapping from $\R$ to $(0,1)$.

% Denote the distribution of $\x,y$ as $\mathscr{P}_{x,y}$.
%\begin{definition}\label{def:nn}
%\textbf{Non-linear regression.} More generally, for $\bt^j\in\R^{r\times d}$, the labels $\yij$ are generated as
%\begin{align*}
%    \yij = f^j(\W\xij) + \epsilon,
%\end{align*}
%$\epsilon$ is a zero mean bounded random variable. 
%\end{definition}

When the dimension of the span of parameters is small, \cite{kong2020meta} performs a dimension reduction algorithm to find the low-dimensional subspace that the parameters span. This is done by selecting the top eigenvectors of the covariance estimate of the cross-correlation between input and labels.
\begin{definition}\label{def:mom}
\textbf{Moment estimator of covariance.} We define the covariance estimator as
\begin{equation}\label{eq:hatM}
\begin{split}
\hat\bM &= \mathop{\Scale[1.8]{\sum}}_{j=1}^k \frac{2}{n_j^2}\left[(\sum_{i=1}^{n_j/2} \yij\xij) (\sum_{i=n_j/2+1}^{n_j} \yij\xij)^\top \right.\\
&\quad\quad\quad\quad \left. +  ( \sum_{i=n_j/2+1}^{n_j} \yij\xij) ( \sum_{i=1}^{n_j/2} \yij\xij)^\top\right].
\end{split}
\end{equation}
Define
\begin{align*}
    &\mtx{h}^j(\W): \R^{r\times d} \rightarrow \R^d = \mtx{E}_{\x}[f^j(\W \x)\x]\\
    &\bM := \W^\top\W\left(\frac{1}{k}\sum_{j=1}^k \mtx{h}^j(\W)( \mtx{h}^j(\W))^\top\right)\W^\top\W.
\end{align*}
\end{definition}
We will prove that $\hat\bM$ is a finite sample estimate of $\bM$.
\noindent\textbf{Subspace estimation.} 
To estimate the subspace $\W$, we use rank-$r$ approximation of $\hat \bM$ to retrieve its principal eigenvector subspace. 
Let $\hat\Ub \hat \La \hat \Ub^\top$ be the eigen-decomposition of $\hat \bM$. Denote $\hat \la_j$ as the $j$th eigenvalue of $\hat \La$. Let $\hat\Ub_r$ be the first $r$ columns of $\hat\Ub$, thus the rank-$r$ approximation is $\hat\bM_r = \hat\Ub_r \mathrm{diag}(\hat \la_{1},...,\hat\la_{r}) \hat \Ub_r^\top$. In the next section, we will prove that the range of $\hat \Ub$ is close to the row space of $\W$.

%{\color{blue} No need to put an algorithm here. If we have a bound for downstream task as well we can fill up the whole procedure.}

%Now we propose the algorithm that accomplishes the subspace retrieval process above. 

In Algorithm \ref{algo:1}, the output $\hat\Ub_r$ is the estimator of the task subspace $\W$. $\hat\Ub_r$ is used as a training step for the few-shot learning phase. For the new task, we search for the function $f^*$ that minimizes the population loss. We shall provide an instructive analysis for a general class of functional family and loss.
\begin{assumption}\label{ass:F}
$\Fc$ is a set of functions satisfying: For any function $f\in\Fc$, any orthonormal matrix $\Qb\in\R^{r\times r}$ and any representation matrix $\Pb\in\R^{r\times d}$, there exists $g \in \Fc$ such that $f(\Pb \x) = g(\Qb\Pb \x)$.
%\begin{enumerate}
%    \item For any function $f\in\Fc$, any orthonormal matrix $\Qb\in\R^{r\times r}$ and any matrix $\Pb\in\R^{r\times d}$, there exists $g \in \Fc$ such that $f(\Pb \x) = g(\Qb\Pb \x)$.
%    \item $\Fc \subseteq \{f\ |\ \log f,~ \log(1-f)\mathrm{~are~}L\mathrm{~Lipschitz}\}\bigcap \{f\ |\ 0<f(x)<1,\forall x\in\R^r\}$.
%\end{enumerate}
\end{assumption}
\ys{In this assumption, we basically mean that the $\Fc$ is invariant with orthonormal rotation $\Qb$. In other words, we only need to find the $r$ dimensional row space $\mathrm{row}(\Pb)$ (to project the features onto it as a low dimensional representation), and don't worry the exact matrix $\Pb$ itself.

Let us introduce population risk $\cL$ and empirical risk $\cL_e$ based on any single loss function between model prediction and true label.}
\begin{align*}
&\cL(f;\Pb) = \mtx{E}_{\Pxy}\cLoss(f(\Pb \x),y)\\
&\cL_e(f;\Pb) = \frac{1}{n}\sum_{i=1}^n\cLoss(f(\Pb \x_i),y_i).
\end{align*}
We make the following assumption on the population risk.
\begin{assumption}\label{ass:L}
Suppose population loss $\cL$ and empirical loss $\cL_e$ satisfy the following assumptions:
\begin{enumerate}[itemsep=-0.1em,topsep=0.3em]
    %\item $\cL$ has the form $\cL(f;\Pb) = \mtx{E}_{\Pxy}\cL(f(\Pb \x),y)$.
    %$\cL_e$ has the form $\cL_e(f;\Pb) = \frac{1}{n}\sum_{i=1}^n\cL(f(\Pb \x_i),y_i)$.
    \item $\cL$ is $L$ Lipschitz in $\Pb \x$.
    \item $\min_{\Pb}\cL(f;\Pb) = \cL(f;\W)$.
\end{enumerate}
\end{assumption}
\noindent\textbf{Example:} Suppose $f$ is an $L$-Lipschitz function with range in $(0,1)$ \ys{and the true labels $y_i$ are from $\{0,1\}$.} The cross entropy function satisfies the assumptions.
\begin{align}
    \cLCE(f;\Pb)=-\mtx{E}_{\Pxy} (&y\log f(\Pb \x)\label{eq:cLCE}\\
   &+ (1-y)\log (1 - f(\Pb \x))) \nonumber.\\
   \cL_e(f;\Pb)=-\frac{1}{n}\sum_{i=1}^n (&y_i\log(f(\Pb \x_i))\label{eq:cLCE finite theta}\\
    &+ (1-y_i)\log (1 - f(\Pb \x_i))).\nonumber 
   %\cL_e(\bt;\Pb)=-\frac{1}{n}\sum_{i=1}^n (&y_i\log(\phi (\bt^\top\Pb \x_i))\label{eq:cLCE finite theta}\\
    %&+ (1-y_i)\log (1 - \phi(\bt^\top\Pb \x_i))).\nonumber 
\end{align}
\ys{With an abuse of notation, we can define the loss with respect to parameterization of the function $f$. For example, if we use the model in Def. \ref{def:GLM}, then we can write the empirical loss as 
\begin{align}
    \cL_e(\bt;\Pb)=-\frac{1}{n}\sum_{i=1}^n (&y_i\log(\phi (\bt^\top\Pb \x_i))\\%\label{eq:cLCE finite theta}\\
    &+ (1-y_i)\log (1 - \phi(\bt^\top\Pb \x_i))).\nonumber
\end{align}
}

% the cross-entropy loss, which is usually employed for classification tasks.
\iffalse
\so{Currently, cross-entropy comes out of nowhere. For using cross-ent, label has to be discrete. It cannot be an arbitrary nonlinear function. Overall, we need to do one of the followings:\\
 Option 1: Do the analysis for a Lipschitz loss function for which cross-entropy is a special case.\\
 Option 2: Let us just say that for few-shot phase, we focus on classification and study cross-entropy. Clearly state label is discrete 0-1.}
 \fi
\begin{definition}\label{def: test}
\textbf{Few-shot classification (Population).} In the few-shot learning phase, suppose $\x,y\sim\Pxy$ satisfy $\mtx{E}[y~|~\x] = f^*(\W\x)$. Let $\Fc$ be a family of functions as the search space for few-shot learning model. 
Let Assumptions \ref{ass:F} and \ref{ass:L} hold. We search for the solution induced by $\hat{\Ub}_r$ by
\begin{align}
    \hat f &= \argmin_{f\in\Fc}\ \cLCE(f;\hat{\Ub}_r^\top)\label{eq:fstar}
\end{align}
%$\cLCE: \Fc\times \R^{r\times d}\rightarrow \R$ be population cross-entropy loss, defined as 
%\begin{align}
%    \cLCE(f;\Pb)=-\mtx{E}_{\Pxy} (&y\log f(\Pb \x)\label{eq:cLCE}\\
%   &+ (1-y)\log (1 - f(\Pb \x))) \nonumber.
%\end{align}

\iffalse
\so{Following my comment above, for cross-ent, clearly $0\leq f\leq 1$. This has to be specified in the definition.}
\fi

%If $\Pb$ is known, we use $\cLCE(f(\Pb x))$ as an equivalent way of representing $\cLCE(f;\Pb)$.
\end{definition}
Observe that, without representation learning, one has to search for both $f$ and $\W$. However with representation learning, we use $\hat{\Ub}_r^\top$ as the representation matrix and only search for $f$.
\begin{remark}\label{rem:logistic}
For the GLM Definition \ref{def:GLM}, we can choose $\Fc$ to be the $\ell_2$ norm constrained functions for some $a\leq\infty$% the set of functions $f$ satisfying
% {\color{red} YS: I want to say $\Fc = \phi\circ\bt\circ(\mathrm{variable})$}
\begin{align}
    \Fc = \{\x\rightarrow \phi(\bt^T\x)~\big|~\|\bt\|_2 \le a,~\bt\in\R^r\}, \label{eq:Fcbin}
\end{align}
Let the new data be generated with $f^*(\W\x) = \phi({\bt^*}^\top\W\x)$ for some ground-truth parameter $\bt^*$. We use $\cLCE(\bt;\Pb)$ to denote the cross-entropy loss in this setting. $\hat f$ (parameterized by $\hat{\bt}$) is given by
\begin{align}
    \hat \bt &= \argmin_{\bt}\ \cLCE(\bt;\hat \Ub_r^\top ),\mbox{~such~that~} \|\bt\| \le a. \label{eq:fstarbin}
\end{align}
\end{remark}

\iffalse
\begin{definition}\label{def: test finite}
\textbf{Finite sample few-shot learning.} In the few-shot learning phase, there are in total $n$ samples, and suppose 
$\x_i\in\R^d$ are generated independently, following standard normal distribution, $y_i\in\R$, for $i=1,...,n$. $\x_i$, $y_i$ satisfy $\mtx{E}(y_i|\x_i) = f^*(\W\x_i)$. Lef $\Fc$ be a family of functions as the search space for few-shot activation function. 
Let $\cL_e$ be empirical loss, satisfying Assumptions \ref{ass:F} and \ref{ass:L}.

%defined as 
%\begin{align}
%    &\cL_e(f;\Pb) : \Fc\times \R^{r\times d}\rightarrow \R= \notag\\
%    &-\frac{1}{n}\sum_{i=1}^n (y_i\log f(\Pb \x_i) + (1-y_i)\log (1 - f(\Pb \x_i))) \label{eq:cLCE finite}
%\end{align}
We search for the solution by
\begin{align}
    \hat f_e &= \argmin_{f\in\Fc}\ \cL_e(f;\Ub_r^\top)\label{eq:fstar finite}
\end{align}

\end{definition}
\fi
% and suppose $\x_i\in\R^d$ are generated independently, following standard normal distribution, $y_i\in\R$, for $i=1,...,n$.
\begin{definition}\label{def: test finite}
\textbf{Few-shot learning (Finite sample GLM).} Suppose there are $n$ samples for new task $(\x_i,y_i)_{i=1}^n$ and $(\x_i,y_i)$ satisfies $\mtx{E}(y_i|\x_i) = f^*(\W\x_i)$. 
%Let $\Fc$ be a family of functions for few-shot learning phase. 
Let $\cL_e$ be empirical loss, satisfying Assumptions \ref{ass:F} and \ref{ass:L}.
%We consider the setup of Remark \ref{rem:logistic}, where $\phi$ is the logistic function. 
%: \Fc\times \R^{r\times d}\rightarrow \R= \notag\\
%\begin{align}
%    \cL_e(\bt;\Pb)=-\frac{1}{n}\sum_{i=1}^n (&y_i\log(\phi (\bt^\top\Pb \x_i))\label{eq:cLCE finite theta}\\
%    &+ (1-y_i)\log (1 - \phi(\bt^\top\Pb \x_i))).\nonumber 
%\end{align}
Given norm constraint $a\leq \infty$, the empirical risk minimizer (ERM) is defined as
\begin{align}
    \hat \bt_e &= \argmin_{\bt}\ \cL_e(\bt;\hat\Ub_r^\top)\quad\text{such~that}\quad\|\bt\|_2\le a.\label{eq:fstar finite theta}
\end{align}
\end{definition}

\setlength{\textfloatsep}{.5em}
\begin{algorithm}[t!]
\caption{Meta-training and Few-shot Learning}\label{algo:1}
\begin{algorithmic}
% with $k$ tasks, the size of data of the $j$-th task has size $n_j$. Dimension $d$. Effective dimension $r$. Set of functions $\Fc$.
\REQUIRE {Dataset $\mathcal{S}$, representation size $r$, function space $\Fc$}
\STATE{Compute $\hat \bM$ via method-of-moments \eqref{eq:hatM}.}
\STATE{Rank $r$ approximation:\\
$\quad\hat\bM_r \leftarrow \hat\Ub_r \mathrm{diag}(\hat \La_{1,1},...,\hat\La_{r,r}) \hat \Ub_r^\top$.}
\STATE{Either~$\hat f \leftarrow \argmin_{f\in\Fc}\ \cLCE(f;\Ub_r^\top) $.
%$\cLCE$ is defined as \eqref{eq:cLCE}.
}
\STATE{Or~~~~~~$\hat f_e \leftarrow \argmin_{f\in\Fc}\ \cL_e(f;\Ub_r^\top) $.
%$\cL_e$ is defined as \eqref{eq:cLCE finite theta}
}
\RETURN{$\hat\Ub_r$ and $\hat f$ or $\hat f_e$.}
\end{algorithmic}
\end{algorithm}%\vspace{-10pt}

\vspace{-0.6cm}
\section{Main Results}
% {\color{red} YS: Commented some text for space limitation}
\vspace{-.1cm}
In this section, we shall establish error bounds for Algorithm \ref{algo:1}. This involves three parts. Theorem \ref{thm:hatM} establishes the quality of the moment estimator $\hat\bM$. %The estimator is then used for few-shot learning, and 
Theorem \ref{thm:down} upper bounds the population cross-entropy risk of $\hat f$ in the few-shot learning stage. Theorem \ref{thm:rade} upper bounds the population risk of the ERM estimator $\hat f_e$, which is learned from finite data.

\subsection{Results on Meta-training}
\begin{lemma}\label{lem:M exp}
$\bM,\hat{\bM}$ satisfies the following. (a) $\text{rank}(\bM)\leq r$. (b) $\text{range-space}(\bM)\subset \text{row-space}(\W)$. (c) $\mtx{E}[\hat{\M}]=\bM$. 
\end{lemma}
In words, $\bM$ returns a consistent estimate of the representation space in the sense that its range is guaranteed to be the subspace of the representation. Observe that to fully recover representation, $\bM$ should contain a diverse set of tasks that can \emph{cover} the representation subspace. For GLM, one needs at least $k\geq r$ tasks to ensure range of $\bM$ is equal to the row-space of $\W$. Additionally, $\hat{\M}$ estimator is also consistent. %All missing proofs can be found in \cite{paperurl}.

% notation (any well-conditioned can be accomplished by estimating the covariance of the inputs and then whitening the data. Such normalization procedures are common e.g.~batch normalization. 
We next present the error on the estimator $\hat \bM$. This theorem applies to standard normal data $\x\sim\Nn(0,\Iden)$. %This is mostly used for notational convenience. In machine learning subGaussian distributions often behave very similar in sufficiently high-dimensions \cite{oymak2018universality,abbasi2019universality}. We leave such generalizations to more general distributions as a future work. 
While this may initially seem restrictive, we remark that identity covariance is mostly used for notational convenience. Additionally, in similar spirit to Central Limit Theorem, machine learning and signal processing algorithms often exhibit distributional universality: For instance, subgaussian distributions often behave very similar or even identical to gaussian distributions in sufficiently high-dimensions \cite{oymak2018universality,abbasi2019universality}. We leave such generalizations to more general distributions as a future work. 
\begin{theorem}[Moment Estimator]\label{thm:hatM}
Suppose the data is generated as in Def.~\ref{def:task}, $n_j\geq N$ for all $j$ and $\xij\distas\Nn(0,\Iden)$. Suppose for some $\sigma>0$ and for all tasks, the label-input product $y\x$ is a subGaussian random vector with covariance upper bounded by $\| \mathbf{Cov}(y \x) \| \le \sigma^2$. (These conditions hold when $|f^j(\x)|<\sigma$.) Let $\delta\in(0,1)$ and $\epsilon\in(0,1)$. Then there exists a constant $c>0$ such that if
\begin{align*}
    k \ge \frac{cd}{N}\log^2(\frac{kd}{\delta}) \max\{\frac{1}{\epsilon^2}, \frac{1}{\epsilon} \log(\frac{kd}{\delta})\},
\end{align*}
$\|\hat \bM - \bM\| \le \epsilon\sigma^2$ with probability at least $1-\delta$.
\end{theorem}
%In essense, this shows that subspace $\M$ can be estimated with $O(kd) samples$
\iffalse
Theorem \ref{thm:hatM} proposes an upper bound for the estimation error of $\bM$. The subspace spanned by $\hat \bM$ is used for few-shot learning. In the context of binary classification, we will show how this error affects the cross entropy loss for the few-shot learning stage. 
\fi
%Downstream task

Recall that $\hat \bM=\hat \Ub \hat\La \hat\Ub^\top$ and $\hat \Ub_r$ is the first $r$ columns of $\hat \Ub$. Denote the estimate of $\W$ via $\hat \W$ given by adjusting $\hat\Ub_r$
\begin{align}
    \hat \W &= (\hat \Ub_r \hat \Qb)^\top,\
    \hat \Qb = \mathop{\mathrm{argmin}}_{\Qb\in\R^{r\times r}, \Qb\Qb^\top = I}\ \|\hat \Ub_r \Qb - \W^\top\|\label{eq:def W}
\end{align}
With the definition of $\hat\W$, $\|\hat \W - \W\|$ defines a distance between the row space of $\W$ and the column space of $\hat \Ub_r$. If the span of the two subspaces are the same, then there exists an orthonormal matrix $\Qb$ such that $\hat \Ub_r \Qb = \W^\top$.

The previous lemma builds upon the assumption that $\|\hat \W - \W\|$ is small. 
%We quote the following lemma first. 
In Theorem \ref{thm:hatM}, we have got $\|\hat \bM - \bM\| \le \epsilon\sigma^2$, then with the extra assumption that $\bM$ is rank $r$, we have the following result from \cite{davis1970rotation}.
\begin{lemma}
If $\|\hat \bM - \bM\| \le \epsilon\sigma^2$, $\lambda_r(\bM) > \epsilon\sigma^2$, then \[\|\hat \W - \W\| \le \epsilon\sigma^2(\lambda_r(\bM) -\epsilon\sigma^2)^{-1}.
%\le \frac{\|\hat \bM - \bM\|}{\lambda_r(\bM) - \|\hat \bM - \bM\|} \le \frac{\epsilon\sigma^2}{\lambda_r(\bM) -\epsilon\sigma^2}.
\]
\end{lemma}

\subsection{Results on Few-shot Learning}
\vspace{-0.1cm}
In the next step, we will use $\hat\Ub_r$ for few shot learning and find $\hat f$ that minimize the population loss. 
%Recall that the search space for $\hat f$ is $\Fc$. For binary classification, we assume 
%\begin{enumerate}
%Only row-space (representation) matters
%    \item \textbf{Rotation invariance:} For any function $f\in\Fc$, any orthonormal matrix $\Qb\in\R^{r\times r}$ and any matrix $\Pb\in\R^{r\times d}$, there exists $g \in \Fc$ such that $f(\Pb \x) = g(\Qb\Pb \x)$.
%    \item \textbf{Lipschitz:} $\Fc \subseteq \{f\ |\ \log f,~ \log(1-f)\mathrm{~are~}L\mathrm{~Lipschitz}\}$ $~\bigcap~ \{f\ |\ 0<f(x)<1,\forall x\in\R^r\}$.
%\end{enumerate}

\begin{theorem}\label{thm:down}
    Let Assumptions \ref{ass:F} and \ref{ass:L} hold. Let $\hat\W$ be the same as in \eqref{eq:def W} and $\hat f$ be the same as in \eqref{eq:fstar}. Then we have 
    \begin{align*}
        \cLCE(\hat f;\hat \Ub_r^\top) - \cLCE(f^*;\W)\lesssim L\sqrt{r}\|\hat \W - \W\|.
    \end{align*}
\end{theorem}
$\cLCE (f^*;\W)$ assumes the knowledge of the true function $f^*$ and the representation $\W$. This shows that the inaccuracy of the moment estimator $\hat M$ costs us $O(L\sqrt{r}\|\hat \W - \W\|)$. 

Theorem \ref{thm:down} bounds the population risk of $\hat f$, when we use $\hat\Ub_r^\top$ as the representation subspace. Next we discuss the population risk of the finite sample solution $\hat f_e$, which should be worse than $\hat f$ due to the limited samples.
%deviating from true data distribution of the new task. Theorem \ref{thm:rade} bounds the risk in terms of the sample size $n$.

\iffalse
For a concrete example, we apply the logistic regression model for \eqref{eq:cLCE finite}, where $f(\vb) = \phi(\bt^\top\vb)$, $\vb \in \R^r$ and $\phi$ is logistic function. Let the ground truth parameter for generating few-shot learning data be $\bt^*$ and we assume $\|\bt^*\| \le a$. In few-shot learning, we train by solving for $\hat \bt_e$ in the set $\{\bt\ |\ \|\bt\| \le a\}$. The objective is {\color{red} can move this to def 2.6}
\begin{align}
    &\cL_e(\bt;\Pb) : \Fc\times \R^{r\times d}\rightarrow \R= \notag\\
    &-\frac{1}{n}\sum_{i=1}^n (y_i\phi (\bt^\top\Pb \x_i) + (1-y_i)\log (1 - \phi(\bt^\top\Pb \x_i))) \label{eq:cLCE finite theta}
\end{align}
The empirical minimizer is defined by
\begin{align}
    \hat \bt_e &= \argmin_{\bt}\ \cL_e(\bt;\Ub_r^\top),\mbox{~such~that~} \|\bt\|\le a.\label{eq:fstar finite theta}
\end{align}
\fi
\begin{theorem}\label{thm:rade}
    Consider the setup in Def.~\ref{def: test finite} with $n$ i.i.d.~examples with ground-truth model $\bt^*$. Solve for $\hat \bt_e$ via \eqref{eq:fstar finite theta}. There exist constants $c>1$, $\delta\in(0,1)$, with probability at least $1 - n^{-c+1} - \delta$, the solution pair $(\hat\bt_e,\hat\Ub_r)$ satisfies
    \begin{align*}
        \quad \mathcal{L}(&\hat\bt_e;\hat\Ub_r^\top) - \mathcal{L} (\bt^*;\W) \\
        &\leq \frac{caL(\sqrt{r}+\log(n)) (1+ \sqrt{\log(1/\delta)})}{\sqrt{n}} + L\sqrt{r}\|\hat \W - \W\|. 
    \end{align*}
\end{theorem}
Note that the first term grows as $\sqrt{r/n}$. This means the amount of data $n$ we request for few-shot learning is $n\approx r$,  as compared to $n\approx d$ if representation learning is not excecuted. 
%If we do not run representation learning before few-shot learning, the error would instead be proportional to $\sqrt{d/n}$ where $d$ is the ambient dimension and we would need $n\approx d$ examples to learn the new task.
% {\color{red} YS:  No math justification for $n\ge d$ here since I only have upper bound but no lower bound, so it's just an intuition}
%In \eqref{eq:fstar}, we solve for the minimizer of population loss function. Now we will study the finite sample case. In few-shot learning phase, suppose we have the dataset $(\x_i, y_i)$, $i=1,...,n$ and the data is generated as Def. \ref{def: test}. Now we consider the 
\vspace{-0.45cm}
\maketitle
\section{Numerical Experiments}\vspace{-5pt}
\label{s:exp}

% {\color{blue} Leave to Ibrahim}

% {\color{blue} I can roughly see what you did, but I'm not sure about a few details. Maybe I can mark down and talk to you on Mon, I'll have enough time to edit after that.}

We generate synthetic datasets with $k$ different tasks and $n$ samples for all tasks. As dimension of the data and dimension of the subspace we choose $d=50$ and $r=5$, respectively. %In all experiments we choose all $n_j$'s to be same i.e.~$n_j=n$. 

We study two different setups. In the first one data is generated according to Def.~\ref{def:GLM}. For the second setup, there is an underlying 3-layer neural network that fits the data. In both setups our only aim is to retrieve subspace representations of the data using Algorithm \ref{algo:1}.

\begin{figure}[t!]
\centering
\begin{subfigure}[htbp!]{0.23\textwidth}
\centering
\includegraphics[width = \linewidth]{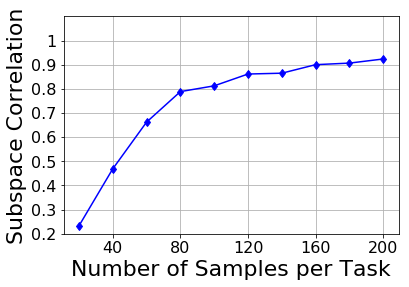}
\subcaption{}\label{fig1a}
\end{subfigure}
\begin{subfigure}[htbp!]{0.23\textwidth}
\centering
\includegraphics[width =\linewidth]{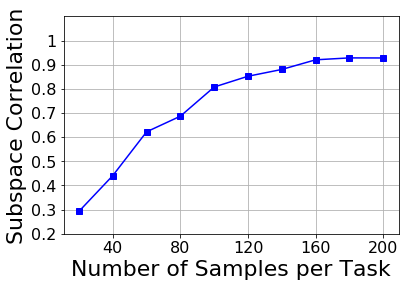}
\subcaption{}\label{fig1b}
\end{subfigure}
\vspace{-0.35cm}
\captionsetup{width=\linewidth}{\caption{Subspace correlations with fixed number of tasks. \hspace{0.4cm}(a) Binary classification (b) Neural network.}  \label{fig1}} 
\end{figure} 

For neural network experiments, we assume that the data are generated from a ground truth neural network which has 3 layers, defined as  
$$ \yij = f^j(\xij) + \epsilon_{i,j} = \mtx{W}_{j3}(\mtx{W}_{j2}(\mtx{W}_{j1}(\mtx{W}\xij))_+)_+ + \epsilon_{i,j}$$ 
where  $\epsilon_{i,j}\sim\Nn(0,1)$ is gaussian noise, $(\cdot)_+$ is the ReLU activation function, $\mtx{W}\in\R^{5\times 50}$ is representation matrix which is same for all $j$'s. The weight matrices $\mtx{W}_{j1},\mtx{W}_{j2}$ and $\vct{W}_{j3}$ are different for each task and they are random gaussian matrices in $\R^{20\times 5},\R^{20\times 20},\R^{20}$ respectively.
\vspace{0cm}
\iffalse
\begin{figure}[htbp!]
\centering
\begin{subfigure}[htbp!]{0.23\textwidth}
\centering
\includegraphics[width = \linewidth]{Paper/Figures_in_Paper/12.png}
\subcaption{}\label{fig2a}
\end{subfigure}
\begin{subfigure}[htbp!]{0.23\textwidth}
\centering
\includegraphics[width =\linewidth]{Paper/Figures_in_Paper/22.png}
\subcaption{}\label{fig2b}
\end{subfigure}
\vspace{-0.2cm}
\captionsetup{width=\linewidth}{\caption{Subspace correlations with fixed number of samples per task and varying number of tasks. (a) Binary classification, (b) Neural network.}  \label{fig2}} 
\end{figure} 
\fi
%{\color{blue} Could you send me the data of the plots? I can work on the figure a bit. It would be better to set ``ylim'' so that the lines span the range of vertical direction. E.g., ylim=[0.4,1] in the first figure. And don't use colors, ICASSP requires grey-scale figures so we can't tell by color.}

In Fig.~\ref{fig1}
%, we apply Algorithm \ref{algo:1} to recover the $r$ dimensional subspace and get $\hat \Ub_r$, and check its distance to the true representation space $\Ub_r$ under different $n$'s and $k$'s. 
we use the subspace correlation as the metric for evaluating the accuracy of subspace recovery, which is defined by  $\frac{\|{\hat{\mtx{U}_r}^\top\mtx{U}_r}\|^2}{\|\mtx{U}_r\|^2}$.
In Fig.~\ref{fig1}, $k=100$ is fixed but $n$'s vary from $20$ to $200$. It can be seen from Fig.~\ref{fig1} that as $nk$ gets bigger, the subspace correlation becomes closer to $1$, which is compatible with Theorem \ref{thm:hatM}

\begin{figure}[t!]
\centering
\begin{subfigure}[htbp!]{0.24\textwidth}
\centering
\includegraphics[width =  \linewidth]{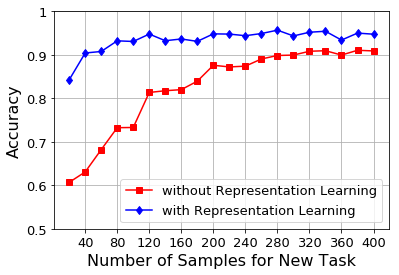}
\end{subfigure}
\hspace{-0.2em}
\begin{subfigure}[htbp!]{0.22\textwidth}
\centering
\includegraphics[width =  \linewidth]{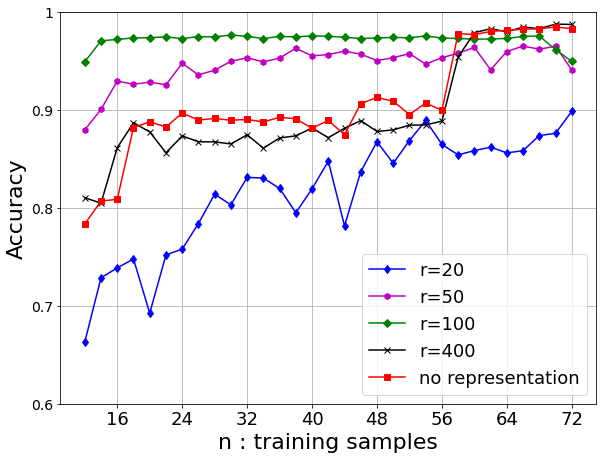}
\end{subfigure}
\captionsetup{width=.45\textwidth}{\caption{(a) Accuracy for downstream task, binary classification (b) Accuracy for downstream the task, MNIST} \label{fig3}}
\end{figure}

In Fig.~\ref{fig3}(a), the downstream task accuracies for binary classification are depicted. For the new task, a new 1-layer neural network without any activation function is trained with and without the retrieved representations of the earlier tasks. We find the parameters of the neural network by minimizing the cross entropy loss via SGD. For this setup, during meta-training, we set $n=50$ for all tasks and $k=2000$, to have almost perfect representation. We evaluate the test error with $1000$ new samples.
%by generating $1000$ samples from the same task, 

If the number of few-shot training samples is small, accuracy improves much faster when we use representation learning. This validates that dimension reduction reduces the degrees-of-freedom for few-shot learning, so the optimal model can be learned with fewer samples. As the sample size grows, the relative benefit of representation is smaller but still noticeable.

In Fig. \ref{fig3}(b), Algorithm \ref{algo:1} is also tested in MNIST dataset. $d=784$ and $r$ is not known. We assume that we have different binary classification tasks among pairs of digits such as 0-1;2-3;0-8;8-4 etc. There exist 15 meta-training tasks (i.e., $k=15$). For each task we have 500 samples in each classes. We choose a different pair of classes as few-shot learning task. We choose binary classification among images of 1 and 9, which is not included in the meta-learning phase. We tune the predicted subspace dimensions and number of samples to get Fig. \ref{fig3}(b).
\iffalse
\begin{figure}[htbp!]
\centering
\begin{subfigure}[htbp!]{0.37\textwidth}
\centering
\includegraphics[width =   \linewidth]{Paper/Figures_in_Paper/alltogether.png}
\end{subfigure}
\captionsetup{width=.45\textwidth}{\caption{Accuracy for downstream the task,MNIST} \label{figlast}}
\end{figure}
\vspace{-0.25cm}
\fi
It can be concluded that when $r=20$, 
%it does not serve a good representation 
The downstream task is not learnt well so we need to expand the subspace. For $r=50$ and $r=100$ subspace learning helps for few-shot learning, as when the number of training samples $n$ is between 8-56 they outperform the case without representation. When $r$ gets closer to $d$, the few-shot sample size has to be large to succeeed due to higher degree of freedom.

\newpage
%\section{REFERENCES}
%\label{sec:refs}
\bibliographystyle{IEEEbib}
\let\oldbibliography\thebibliography
\renewcommand{\thebibliography}[1]{%
  \oldbibliography{#1}%
  \setlength{\itemsep}{-1pt}%
}
\bibliography{refs}

\begin{thebibliography}{10}

\bibitem{caruana1997multitask}
Rich Caruana,
\newblock ``Multitask learning,''
\newblock {\em Machine learning}, vol. 28, no. 1, pp. 41--75, 1997.

\bibitem{baxter2000model}
Jonathan Baxter,
\newblock ``A model of inductive bias learning,''
\newblock {\em Journal of artificial intelligence research}, vol. 12, pp.
  149--198, 2000.

\bibitem{deng2009imagenet}
Jia Deng, Wei Dong, Richard Socher, Li-Jia Li, Kai Li, and Li~Fei-Fei,
\newblock ``Imagenet: A large-scale hierarchical image database,''
\newblock in {\em 2009 IEEE conference on computer vision and pattern
  recognition}. Ieee, 2009, pp. 248--255.

\bibitem{bojar2014findings}
Ond{\v{r}}ej Bojar, Christian Buck, Christian Federmann, Barry Haddow, Philipp
  Koehn, Johannes Leveling, Christof Monz, Pavel Pecina, Matt Post, Herve
  Saint-Amand, et~al.,
\newblock ``Findings of the 2014 workshop on statistical machine translation,''
\newblock in {\em Proceedings of the ninth workshop on statistical machine
  translation}.

\bibitem{finn2017model}
Chelsea Finn, Pieter Abbeel, and Sergey Levine,
\newblock ``Model-agnostic meta-learning for fast adaptation of deep
  networks,''
\newblock in {\em International Conference on Machine Learning}, 2017, pp.
  1126--1135.

\bibitem{schmidhuber1987evolutionary}
J{\"u}rgen Schmidhuber,
\newblock {\em Evolutionary principles in self-referential learning, or on
  learning how to learn: the meta-meta-... hook},
\newblock Ph.D. thesis, Technische Universit{\"a}t M{\"u}nchen, 1987.

\bibitem{thrun2012learning}
Sebastian Thrun and Lorien Pratt,
\newblock {\em Learning to learn},
\newblock Springer Science \& Business Media, 2012.

\bibitem{bengio2013representation}
Yoshua Bengio, Aaron Courville, and Pascal Vincent,
\newblock ``Representation learning: A review and new perspectives,''
\newblock {\em IEEE transactions on pattern analysis and machine intelligence},
  vol. 35, no. 8, pp. 1798--1828, 2013.

\bibitem{hospedales2020meta}
Timothy Hospedales, Antreas Antoniou, Paul Micaelli, and Amos Storkey,
\newblock ``Meta-learning in neural networks: A survey,''
\newblock {\em arXiv preprint arXiv:2004.05439}, 2020.

\bibitem{yosinski2014transferable}
Jason Yosinski, Jeff Clune, Yoshua Bengio, and Hod Lipson,
\newblock ``How transferable are features in deep neural networks?,''
\newblock in {\em Advances in neural information processing systems}, 2014, pp.
  3320--3328.

\bibitem{zhong2016mixed}
Kai Zhong, Prateek Jain, and Inderjit~S Dhillon,
\newblock ``Mixed linear regression with multiple components,''
\newblock in {\em Advances in neural information processing systems}, 2016, pp.
  2190--2198.

\bibitem{li2018learning}
Yuanzhi Li and Yingyu Liang,
\newblock ``Learning mixtures of linear regressions with nearly optimal
  complexity,''
\newblock in {\em Conference On Learning Theory}, 2018, pp. 1125--1144.

\bibitem{chen2020learning}
Sitan Chen, Jerry Li, and Zhao Song,
\newblock ``Learning mixtures of linear regressions in subexponential time via
  fourier moments,''
\newblock in {\em Proceedings of the 52nd Annual ACM SIGACT Symposium on Theory
  of Computing}.

\bibitem{lounici2011oracle}
Karim Lounici, Massimiliano Pontil, Sara Van De~Geer, Alexandre~B Tsybakov,
  et~al.,
\newblock ``Oracle inequalities and optimal inference under group sparsity,''
\newblock {\em The annals of statistics}, vol. 39, no. 4, pp. 2164--2204, 2011.

\bibitem{cavallanti2010linear}
Giovanni Cavallanti, Nicolo Cesa-Bianchi, and Claudio Gentile,
\newblock ``Linear algorithms for online multitask classification,''
\newblock {\em The Journal of Machine Learning Research}, vol. 11, pp.
  2901--2934, 2010.

\bibitem{maurer2016benefit}
Andreas Maurer, Massimiliano Pontil, and Bernardino Romera-Paredes,
\newblock ``The benefit of multitask representation learning,''
\newblock {\em The Journal of Machine Learning Research}, vol. 17, no. 1.

\bibitem{balcan2015efficient}
Maria-Florina Balcan, Avrim Blum, and Santosh Vempala,
\newblock ``Efficient representations for lifelong learning and autoencoding,''
\newblock in {\em Conference on Learning Theory}.

\bibitem{rish2008closed}
Irina Rish, Genady Grabarnik, Guillermo Cecchi, Francisco Pereira, and
  Geoffrey~J Gordon,
\newblock ``Closed-form supervised dimensionality reduction with generalized
  linear models,''
\newblock in {\em Proceedings of the 25th international conference on Machine
  learning}, 2008, pp. 832--839.

\bibitem{argyriou2008convex}
Andreas Argyriou, Theodoros Evgeniou, and Massimiliano Pontil,
\newblock ``Convex multi-task feature learning,''
\newblock {\em Machine learning}, vol. 73, no. 3, pp. 243--272, 2008.

\bibitem{kong2020meta}
Weihao Kong, Raghav Somani, Zhao Song, Sham Kakade, and Sewoong Oh,
\newblock ``Meta-learning for mixed linear regression,''
\newblock {\em arXiv preprint arXiv:2002.08936}, 2020.

\bibitem{kong2020robust}
Weihao Kong, Raghav Somani, Sham Kakade, and Sewoong Oh,
\newblock ``Robust meta-learning for mixed linear regression with small
  batches,''
\newblock {\em arXiv preprint arXiv:2006.09702}, 2020.

\bibitem{du2020few}
Simon~S Du, Wei Hu, Sham~M Kakade, Jason~D Lee, and Qi~Lei,
\newblock ``Few-shot learning via learning the representation, provably,''
\newblock {\em arXiv:2002.09434}, 2020.

\bibitem{tripuraneni2020provable}
Nilesh Tripuraneni, Chi Jin, and Michael~I Jordan,
\newblock ``Provable meta-learning of linear representations,''
\newblock {\em arXiv preprint arXiv:2002.11684}, 2020.

\bibitem{oymak2018universality}
Samet Oymak and Joel~A Tropp,
\newblock ``Universality laws for randomized dimension reduction, with
  applications,''
\newblock {\em Information and Inference: A Journal of the IMA}, vol. 7, no. 3,
  pp. 337--446, 2018.

\bibitem{abbasi2019universality}
Ehsan Abbasi, Fariborz Salehi, and Babak Hassibi,
\newblock ``Universality in learning from linear measurements,''
\newblock in {\em Advances in Neural Information Processing Systems}, 2019, pp.
  12372--12382.

\bibitem{davis1970rotation}
Chandler Davis and William~Morton Kahan,
\newblock ``The rotation of eigenvectors by a perturbation. iii,''
\newblock {\em SIAM Journal on Numerical Analysis}, vol. 7, no. 1, pp. 1--46,
  1970.

\bibitem{jin2019short}
Chi Jin, Praneeth Netrapalli, Rong Ge, Sham~M Kakade, and Michael~I Jordan,
\newblock ``A short note on concentration inequalities for random vectors with
  subgaussian norm,''
\newblock {\em arXiv preprint arXiv:1902.03736}, 2019.

\bibitem{mohri2018foundations}
Mehryar Mohri, Afshin Rostamizadeh, and Ameet Talwalkar,
\newblock {\em Foundations of machine learning},
\newblock MIT press, 2018.

\bibitem{laurent2000adaptive}
Beatrice Laurent and Pascal Massart,
\newblock ``Adaptive estimation of a quadratic functional by model selection,''
\newblock {\em Annals of Statistics}, pp. 1302--1338, 2000.

\end{thebibliography}
\clearpage
\appendix
\end{document}